\documentclass[letterpaper]{article} 
\usepackage{aaai25}  
\usepackage{times}  
\usepackage{helvet}  
\usepackage{courier}  
\usepackage[hyphens]{url}  
\usepackage{graphicx} 
\usepackage{natbib}  
\usepackage{caption} 
\usepackage{algorithm}
\usepackage{algorithmic}

\usepackage{newfloat}
\usepackage{listings}
\DeclareCaptionStyle{ruled}{labelfont=normalfont,labelsep=colon,strut=off} 
\floatstyle{ruled}
\newfloat{listing}{tb}{lst}{}
\floatname{listing}{Listing}
\usepackage{booktabs}
\usepackage{stmaryrd}
\usepackage[accsupp]{axessibility}
\usepackage{mathpartir}
\usepackage{comment}
\usepackage{mathrsfs}
\usepackage{pifont}
\usepackage{enumitem}
\usepackage{placeins}
\usepackage{xcolor}
\newcommand{\algoname}{\textsc{SYNAPSE}}

\newcommand{\tick}{\ding{51}}
\newcommand{\cros}{\ding{55}}

\newcommand{\anon}[1]{#1}

\newcommand{\program}{\ensuremath{\pi}}

\newcommand{\observationspace}{\ensuremath{O}}

\newcommand{\queryspace}{\ensuremath{Q}}

\newcommand{\preftask}{\ensuremath{\mathcal{T}}}
\newcommand{\prefspace}{\ensuremath{P}}
\newcommand{\conceptlibrary}{\ensuremath{\mathcal{C}}}

\definecolor{rule_label_color}{rgb}{0, 0.3, 0.0}
\definecolor{demo_color}{rgb}{0.6, 0, 0.0}
\definecolor{stx_color}{rgb}{0, 0, 0.8}
\definecolor{cmt_color}{rgb}{0.5, 0.7, 0.3}

\definecolor{grey12}{rgb}{0.98,0.98,0.98}
\definecolor{grey11}{rgb}{0.96,0.96,0.96}
\definecolor{grey10}{rgb}{0.93,0.93,0.93}
\definecolor{grey9}{rgb}{0.9,0.9,0.9}
\definecolor{grey8}{rgb}{0.8,0.8,0.8}
\definecolor{grey7}{rgb}{0.7,0.7,0.7}
\definecolor{grey6}{rgb}{0.6,0.6,0.6}
\definecolor{grey5}{rgb}{0.5,0.5,0.5}
\definecolor{grey4}{rgb}{0.4,0.4,0.4}
\definecolor{grey3}{rgb}{0.3,0.3,0.3}
\definecolor{grey2}{rgb}{0.2,0.2,0.2}

\definecolor{rule_label_color}{rgb}{0, 0.3, 0.0}
\newcommand{\algcmt}[1]{\textcolor{cmt_color}{\emph{\# #1}}}

\RequirePackage{xspace}
\makeatletter
\DeclareRobustCommand\onedot{\futurelet\@let@token\@onedot}
\def\@onedot{\ifx\@let@token.\else.\null\fi\xspace}

\def\eg{\emph{e.g}\onedot} 

\def\ie{\emph{i.e}\onedot}

\def\etc{\emph{etc}\onedot} 
\def\vs{\emph{vs}\onedot}

\makeatother

\title{\algoname: SYmbolic Neural-Aided Preference Synthesis Engine}
\author{
    Sadanand Modak\textsuperscript{\rm 1}\thanks{Corresponding author.},
    Noah Patton\textsuperscript{\rm 1},
    Isil Dillig\textsuperscript{\rm 1},
    Joydeep Biswas\textsuperscript{\rm 1}
}
\affiliations{
    \textsuperscript{\rm 1}Department of Computer Science, The University of Texas at Austin\\
    \{smodak11, npatt, isil, joydeepb\}@cs.utexas.edu
}

\begin{document}

\maketitle

\begin{abstract}
This paper addresses the problem of \emph{preference learning}, which aims to align robot behaviors through learning user-specific preferences (\eg ``good pull-over location'') from visual demonstrations. Despite its similarity to learning \emph{factual} concepts (\eg ``red door''), preference learning is a fundamentally harder problem due to its subjective nature and the paucity of person-specific training data. We address this problem using a novel framework called \algoname, which is a neuro-symbolic approach designed to efficiently learn preferential concepts from limited data. \algoname\ represents preferences as neuro-symbolic programs -- facilitating inspection of individual parts for alignment -- in a domain-specific language (DSL) that operates over images and leverages a novel combination of visual parsing, large language models, and program synthesis to learn programs representing individual preferences. We perform extensive evaluations on various preferential concepts as well as user case studies demonstrating its ability to align well with dissimilar user preferences. Our method significantly outperforms baselines, especially when it comes to out-of-distribution generalization. We show the importance of the design choices in the framework through multiple ablation studies. Code, additional results and supplementary material can be found on the website: \texttt{https://amrl.cs.utexas.edu/synapse}
\end{abstract}
\section{Introduction} \label{sec:intro}
\begin{figure*}[t]
    \centering
    \includegraphics[width=1.0\textwidth]{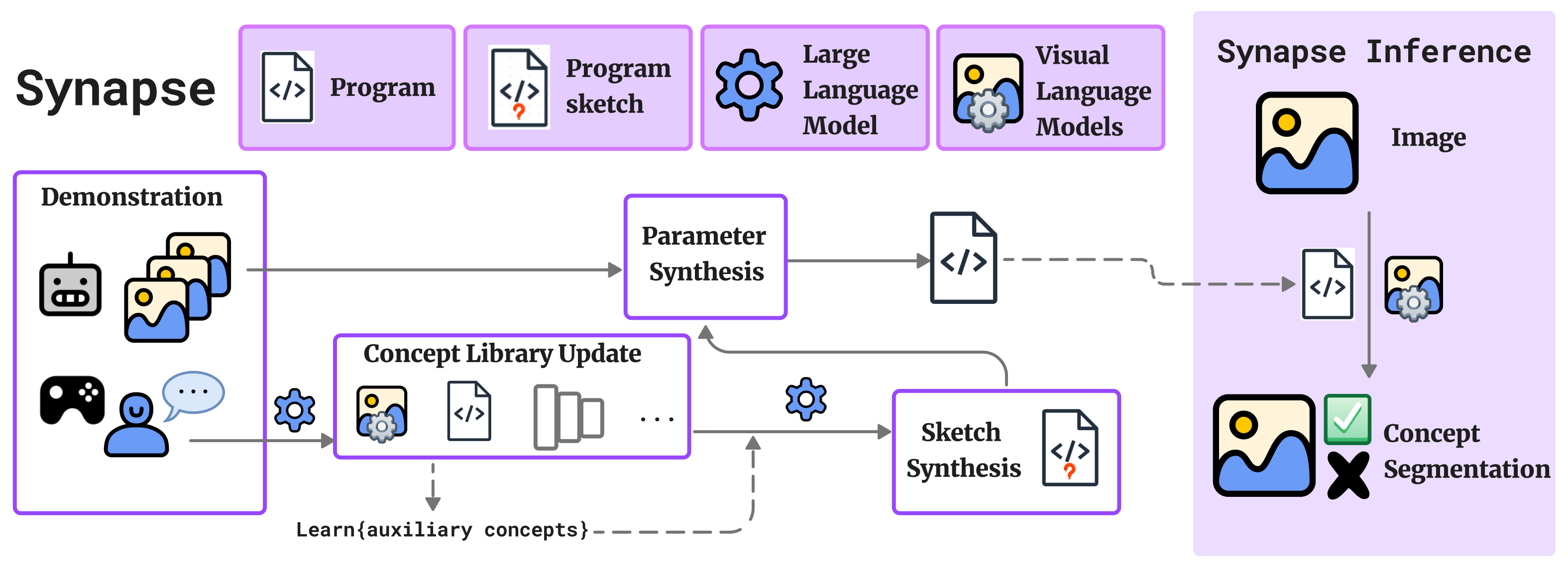}
    \caption{\textbf{Overview.} Human preferences have both \emph{qualitative} and \emph{quantitative} aspects. \algoname\ first learns the necessary predicates (\emph{a.k.a.} auxiliary concepts) needed to represent the preference from the NL input. It then synthesizes a program sketch which likely has some quantitative holes. This sketch represents the preference \emph{qualitatively}. Finally, the holes are filled up by an optimization process that uses the physical demonstration data, thereby capturing the \emph{quantitative} part of the preference.}
    \label{fig:framework}
\end{figure*}
Imagine trying to come up with a definition of \emph{``a good taxi drop-off location''}. One person may consider a spot to be a good drop-off location depending on whether it is close to the door of a building, while someone else might want it in the shade. Such concepts vary from person to person and inherently depend on their preferences. We call them \emph{preferential concepts}, and we are interested in the problem of \emph{preference learning} from visual input. Learning preferences is important because we want robots, and systems in general, that are customizable and can adapt to end-users (\eg home robots). This problem is related to the task of visual concept learning, which focuses on learning factual concepts such as \emph{having the color red} or \emph{being to the left of another object}~\cite{JCL2017, NSCL2019, VCML2019, DCL2021, FALCON2022, NS3D2023, Voyager2023, LEFT2023}. All such prior work assumes there is a \emph{ground-truth} for the concept, \ie the definition of the concept does not differ among people, and as a consequence, sufficiently many examples are available and can be objectively evaluated. We refer to such concepts as \emph{factual concepts}. While most prior work that learns visual concepts exploits the availability of large datasets such as CLEVR~\cite{johnson2017CLEVRvqa}, those methods cannot be applied to  preference learning  because it is a data-impoverished setting by its very nature: a single individual can only provide limited training data. Reinforcement learning-based preference learning, where human preferences are represented as neural networks or latent reward models~\cite{rlhf, drpl29, drpl31}, also suffers from this limitation, and is thus applicable only to learning consistent preferences across a large population, rather than individual preferences. Furthermore, because preferences are inherently individual, they can depend on entirely different (auxiliary lower-level) concepts, such as in the drop-off location example above (\ie based on \emph{proximity to door} as opposed to \emph{being in the shade}). This requires learning novel visual concepts in a \emph{hierarchical} manner, \ie first learning the auxiliary concepts and then the ego concept. Lastly, coming up with a complete definition of a preferential concept at once is itself a hard problem: it is much easier for someone to \emph{incrementally} demonstrate examples one-at-a-time that satisfy their intuition as humans tend to \textit{build} their notion of a preferential concept over time. Thus, preference learning calls for an approach that can handle \emph{incremental learning} from visual demonstrations, since the robot might not have access to a large number of samples at once.

To address these challenges, we present \algoname, a novel framework that learns human preferences in a data-efficient manner. In contrast to prior preference learning approaches which take in weak reward signals to learn preferences, we use a more direct form of a signal, which consists of a robot-demonstration and a natural language (NL) explanation for the preference. We use NL input to identify new concepts to be learned as well as how to compose them, thus representing the preference qualitatively. However, in addition to learning new concepts or composing existing ones, preferences also have a quantitative aspect. For instance, to be a good drop-off spot, it should be close to a door, but exactly \emph{how close} is a personal preference. This is where the demonstrations come into play and allow us to infer quantitative aspects of the preference that are hard to capture via natural language alone. Finally, to allow \emph{incremental}, \emph{sample-efficient} learning, \algoname\ expresses preferential concepts as programs in  a \emph{neuro-symbolic} domain specific language (DSL) operating over images, and learns these programs based on demonstrations. Such a programmatic representation also facilitates \emph{life-long learning}, allowing incremental changes to the learnt program as new demonstrations arrive.

Figure \ref{fig:framework} shows a schematic of our proposed \algoname\ framework. Given a user demonstration (\ie the physical demonstration and NL input), the general workflow of \algoname\ has three main components. First, \algoname\ leverages the user's NL explanation, along with \algoname's existing \emph{concept library} (a collection of auxiliary learned concepts so far), to ground the \emph{concepts} needed to represent the user's preference. If the NL explanation contains concepts that are not part of \algoname's existing concept library, \algoname\ may query the user for additional demonstrations of the auxiliary concept, which are then used to update \algoname's concept library. Once the library contains all required concepts, \algoname\ uses the NL explanation to generate a \emph{program sketch} which is a program in our DSL with missing values (holes) for numeric parameters. Finally, \algoname\ uses constrained optimization techniques (based on \emph{maximum satisfiability}~\cite{LDIPS2021}) to find values of the numeric parameters that are maximally consistent with the user's physical demonstrations.

In what follows, we first describe the state-of-the-art in the field (Section~\ref{sec:relatedwork}) followed by details on the proposed framework (Section~\ref{sec:method}). This paper focuses mainly on mobility related preferences concerning navigation of robots and autonomous vehicles. However, to show that \algoname\ works equally well for other concepts as well, we apply it to a well-studied preferential task in robot manipulation -- tabletop object rearrangement~\cite{LLMGROP2023}. Section~\ref{sec:evaluation} presents our extensive evaluation on these domains, which includes user-studies as well as ablations. Finally, we conclude with a discussion on the opportunities for further work (Section~\ref{sec:discussion}). In summary, this paper contributes:
\begin{enumerate}
    \item \algoname, a neuro-symbolic framework to learn and evaluate preferences
    \item a novel method for hierarchical lifelong learning from visual demonstrations and NL explanation
    \item a comprehensive experimental evaluation of the proposed approach showing generalization to various domains
\end{enumerate}

\section{Related work} \label{sec:relatedwork}
\algoname\ positions itself in the larger field of concept learning and visual question answering (VQA). While there exist reinforcement learning based methods for preference learning, most of them fall in the imitation learning setting, where human preferences are represented via  neural policies~\cite{drpl28, drpl29} or latent reward models~\cite{drpl31, drpl32, drpl34}. Further, they do not deal with natural language, but rather take some form of weak preference signal as input. In the following discussion, we focus on work that is most closely-related to \algoname.

\subsubsection{Language Model Programs (LMPs).} Generating executable programs from natural language is not a new idea. Many earlier works~\cite{JCL2017, NSVQA2018, NSCL2019, DCL2021, FALCON2022} use custom semantic parsers to perform specific tasks. However, with the advent of Large Language Models (LLMs), LMPs have gained significant attention~\cite{Codebotler2023, VisProg2023, ViperGPT2023, LLMP2023} due to the extensive knowledge that these foundation models possess. Code-as-Policies~\cite{CodeAsPolicies2023} pioneered the effort in this direction and demonstrated that LLMs can generate simple Python programs through recursive prompting for tasks ranging from drawing shapes to tabletop manipulation. While this can work for simpler and obvious tasks, it cannot be employed for learning preferences where people can associate different meanings to certain predicates, \eg \texttt{is\_close} might quantitatively differ between two individuals. As we describe later, \algoname\ tries to remedy this by actively querying the human demonstrator for the auxiliary predicate meaning.


\subsubsection{Neuro-symbolic concept learning.} Neuro-symbolic approaches~\cite{VCML2019, FALCON2022, RCL2022, BPNS2022, NS3D2023} couple the interpretability of rule-based symbolic AI with the strength of neural networks. One work~\cite{JCL2017} uses a trained semantic parser to first extract useful feature definitions from a few statements describing the concept, which are then evaluated for each datapoint to build feature vectors on which standard classification can be done. NS-VQA~\cite{NSVQA2018} is another approach that uses a separately trained visual parser to generate a structured representation of objects in the image and a semantic parser trained to parse the question into a predefined DSL format, which is followed by Pythonic program execution. NS-CL~\cite{NSCL2019} uses the same framework as NS-VQA, but instead of answering questions given the trained modules, it represents concepts as neural operators and tries to learn them given the question-answer pairs. However, most of these concept learning methods are data-hungry which makes them unfit for learning preferences.

\subsubsection{Program synthesis.} There is a rich literature on synthesizing programs from user-provided specifications in the programming languages community ~\cite{synthesis-survey,prolex,flashfill,LDIPS2021}. There also has been work on synthesizing LTL formulas directly from natural language~\cite{ltl1,ltl2}. From the viewpoint of visual reasoning and concept learning, \cite{VDP2019} tries to solve Visual Discrimination Puzzles (VDP) by synthesizing a discriminator expressed in first-order logic by performing a full-blown discrete search. However, this can quickly become inefficient as problem size scales. To tackle this, \algoname\ uses natural language informed sketch generation and performs synthesis over the space of parameters.
\begin{figure*}[t]
\centering
\includegraphics[width=0.9\textwidth]{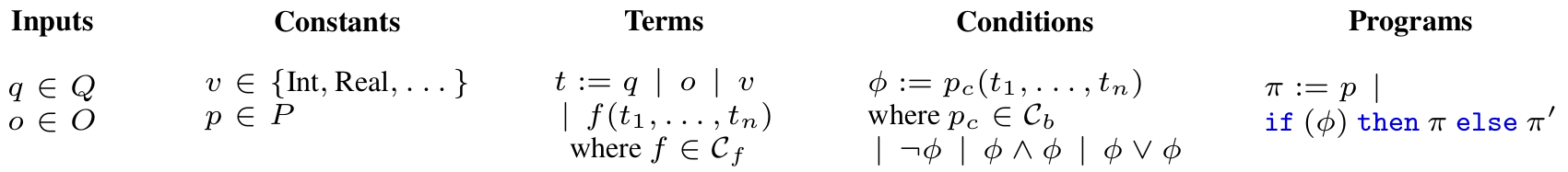}
\caption{\textbf{\algoname\ neuro-symbolic DSL.} Representing preference evaluator $\pi$ parametrized over concept library $\conceptlibrary$}
\label{fig:dsl}
\end{figure*}

\section{Method} \label{sec:method}
We define a \emph{preference task} $\preftask{}:=\langle \observationspace, \queryspace, \prefspace\rangle$ as a tuple consisting of an observation space \observationspace{}, a query space \queryspace{}, and a preference space \prefspace{}. A preference evaluator $\program : \observationspace \times \queryspace \rightarrow \prefspace$ accepts an observation and a query, and returns a preference value $p \in \prefspace$. The goal of \emph{preference learning} is to synthesize a suitable evaluator $\program$ that accurately predicts a person's preferences. Since we focus on visual preferences in this paper, $\observationspace$ is the space of RGB images, $\queryspace$ is pixel/location queries, and $\prefspace$ is a segmentation mask over the image. 

\subsection{Representing Preferences} 
As stated earlier, a distinguishing feature of preference learning is that it has to be performed using \emph{small} amounts of training data. To enable sample-efficient learning, \algoname\ represents the preference evaluator $\program$ as a neuro-symbolic program in a DSL and synthesizes $\program$ from a small number of user demonstrations, where each demonstration includes a robot trajectory\footnote{this applies to the mobility concepts; for the tabletop rearrangement task it is just a single image demonstration} (\ie sequence of images) along with an NL explanation for the user's preference. The DSL, as shown in Figure~\ref{fig:dsl}, is parameterized over a \emph{concept library} $\conceptlibrary$, which includes both boolean predicates $\conceptlibrary_b$ and non-boolean functions $\conceptlibrary_f$, built-in operators (\eg $+, \leq, \ldots$), pre-trained neural models (\eg zero-shot visual language models (VLMs)), as well as previously learned concepts and functions (expressed in the same DSL). At a high level, a program $\program$ in this DSL consists of (nested) if-then-else statements and is therefore conceptually similar to a decision tree. Each leaf of this decision tree is a preference (\eg \emph{good, neutral, bad}) drawn from the preference space $P$, which is assumed to be a finite set. Internal nodes of the decision tree are neuro-symbolic conditions $\phi$, which include boolean combinations of predicates of the form $p_c(t_1, \ldots, t_n)$ where $p_c \in \conceptlibrary_b$ and each $t_i$ is a neuro-symbolic term, that could be an operator (\eg $\leq$), a neural module output, or a previously-learned concept (\eg \texttt{is\_close}).

\subsection{Learning Preferences}
Algorithm \ref{alg:learn} summarizes the learning algorithm for synthesizing preference evaluator $\pi$ from a set of demonstrations. As \algoname\ is meant to be used in a lifelong learning setting, we present it as an incremental algorithm that takes one new demonstration $d_{new}$ at each invocation and returns an updated preference evaluation function. As mentioned earlier, we represent each demonstration $d$ as a pair $(t, e)$ where $t$ is a physical demonstration consisting of a sequence of images from a robot trajectory and $e$ is the user's NL explanation for their preference. Given a demonstration $d$, we write $d.t$ and $d.e$ to denote its physical demonstration and NL component respectively. In addition to the new demonstration $d_{new}$, \texttt{Learn} takes three additional arguments, namely the previous set of demonstrations $\mathcal{D}$, the previously learned preference evaluator $\pi_o$ (\texttt{None} for the first invocation), and the current concept library $\conceptlibrary$, which is initialized to contain only a set of built-in concepts (\ie set of basic mathematical operations, camera homography, and pre-trained neural modules). \texttt{Learn} uses the old program $\pi_o$ to bootstrap the learning process, and the previous demonstrations are required to ensure that the updated program is consistent with \emph{all} demonstrations provided thus far. At a high level, the learning procedure consists of three steps which will be explained in the remainder of this section.

\begin{algorithm}[tb]
\caption{\textbf{The SYNAPSE learning framework}}\label{alg:learn}
\textbf{Input}: a set of previously seen demonstrations $\mathcal{D}$, the new demonstration $d_{new}$, previous sketch $\hat{\pi}_o$, and previous $\mathcal{C}$\\
\textbf{Output}: the new demonstrations set $\mathcal{D}'$, a neuro-symbolic preference evaluator $\pi$ parameterized by new concept library $\mathcal{C}'$, new sketch $\hat{\pi}$ used to generate $\pi$
\begin{algorithmic}[1] 
    \STATE $\mathsf{Learn}(\mathcal{D}, d_{new}, \hat{\pi}_o, \mathcal{C})$
    \STATE \ \ \ \ \algcmt{Update concept library with new NL utterance}
    \STATE \ \ \ \ $\mathcal{C}' \gets \mathsf{UpdateConceptLibrary}(d_{new}.e, \mathcal{C})$
    \STATE \ \ \ \ \algcmt{Get the updated sketch from the new\\ \ \ \ \ \ \ demonstration and previous sketch}
    \STATE \ \ \ \ $\hat{\pi} \gets \mathsf{SketchSynth}(d_{new}, \hat{\pi}_o, \mathcal{C}')$
    \STATE \ \ \ \ \algcmt{Fill the holes in $\hat{\pi}$ based on demonstrations $\mathcal{D}'$}
    \STATE \ \ \ \ $\pi \gets \mathsf{ParamSynth}(\hat{\pi}, \mathcal{D} \cup \{ d_{new} \})$
    \STATE \ \ \ \ {\bf return} $\mathcal{D} \cup \{ d_{new} \}, \pi, \mathcal{C}', \hat{\pi}$
\end{algorithmic}
\end{algorithm}

\begin{algorithm}[tb]
\caption{\textbf{Concept library update}}\label{alg:concept_lib_up}
\textbf{Input}: a new NL explanation $e$ and the previous concept library $\mathcal{C}$\\
\textbf{Output}: a new concept library $\mathcal{C}'$
\begin{algorithmic}[1] 
    \STATE $\mathsf{UpdateConceptLibrary}(e, \mathcal{C})$
    \STATE \ \ \ \ $\mathcal{C}' \gets \mathcal{C}$ \algcmt{Initialize new concept library}
    \STATE \ \ \ \ \algcmt{Get new visual groundings from NL and add to $\mathcal{C}'$}
    \STATE \ \ \ \ $g \gets \mathsf{ExtractEntities}(e, \mathcal{C})$
    \STATE \ \ \ \ $\mathcal{C}' \gets \mathcal{C}' \cup g$
    \STATE \ \ \ \ \algcmt{Extract new predicates from $e$}
    \STATE \ \ \ \ $\mathrm{preds} \gets \mathsf{ExtractPredicates}(e, \mathcal{C})$
    \STATE \ \ \ \ \algcmt{Recursively update concepts with user feedback}
    \STATE \ \ \ \ {\bf for} $\mathrm{pred}\ \in \mathrm{preds}\ \textbf{where}\ \mathrm{pred}\ \notin\ \mathcal{C}$
    \STATE \ \ \ \ \ \ \ \ $\mathcal{D} \gets \emptyset\ \ \algcmt{Empty initial demonstration set}$
    \STATE \ \ \ \ \ \ \ \ $\mathcal{C}'' \gets \mathcal{C}'$
    \STATE \ \ \ \ \ \ \ \ $\hat{\pi} \gets \mathrm{None}$
    \STATE \ \ \ \ \ \ \ \ {\bf do}
    \STATE \ \ \ \ \ \ \ \ \ \ \ \ $d \gets \mathsf{QueryUserForDemonstration}(\mathrm{pred})$
    \STATE \ \ \ \ \ \ \ \ \ \ \ \ $\mathcal{D}, \pi, \mathcal{C}'', \hat{\pi} \gets \mathsf{Learn}(\mathcal{D}, d, \hat{\pi}, \mathcal{C}'')$
    \STATE \ \ \ \ \ \ \ \ {\bf while} $d$
    \STATE \ \ \ \ \ \ \ \ $\mathcal{C}' \gets \mathcal{C}''$
    \STATE \ \ \ \ {\bf return} $\mathcal{C}'$
\end{algorithmic}
\end{algorithm}

\begin{algorithm}[tb]
\caption{\textbf{Parameter synthesis}}\label{alg:param_synth}
\textbf{Input}: a program sketch $\hat{\pi}$, a set of demonstrations $\mathcal{D}$\\
\textbf{Output}: a complete program $\pi$
\begin{algorithmic}[1] 
    \STATE $\mathsf{ParamSynth}(\hat{\pi}, \mathcal{D})$
    \STATE \ \ \ \ $\varphi \gets$ true \algcmt{Initialize}
    \STATE \ \ \ \ {\bf for} $d \in D$
    \STATE \ \ \ \ \ \ \ \ \algcmt{Perform partial evaluation on the sketch\\ \ \ \ \ \ \ \ \ \ and demonstration to get a simplified\\ \ \ \ \ \ \ \ \ \ sketch $\hat{\pi}'$ and expected result $r$}
    \STATE \ \ \ \ \ \ \ \ $(\hat{\pi}', r) \gets \mathsf{PartialEval}(\hat{\pi}, d)$
    \STATE \ \ \ \ \ \ \ \ \algcmt{Merge with condition}
    \STATE \ \ \ \ \ \ \ \ $\varphi \gets \varphi \ \land \ [\![\hat{\pi}']\!]^r$
    \STATE \ \ \ \ \ \ \ \algcmt{Include negation for each parameter $\in P$}
    \STATE \ \ \ \ \ \ \ {\bf for} $i \in P$
    \STATE \ \ \ \ \ \ \ \ \ \ \ {\bf if} $(i \neq r)\ \varphi \gets \varphi\ \land  \neg [\![\hat{\pi}']\!]^i$
    \STATE \ \ \ \ \algcmt{Use solver to fill holes over symbolic features}
    \STATE \ \ \ \ $\pi \gets \mathsf{Solver}(\varphi)$
    \STATE \ \ \ \ {\bf return} $\pi$
\end{algorithmic}
\end{algorithm}


\subsubsection{Concept Library Update.} First, we check whether the existing concept library $\conceptlibrary$ is sufficient for successfully learning the desired preference evaluation function by analyzing the user's NL explanation $e$ to extract concepts of interest. We differentiate between two types of concepts: (1) entities (\eg car, door, sidewalk) and (2) predicates (\eg far, near). Because we use an open-vocabulary VLM to find entities of interest in the current observation, new entity concepts do not require interacting with the user. On the other hand, if NL contains new predicates that are not part of the existing concept library, it needs to query the user to provide suitable demonstrations. For instance, if user provides NL as \emph{``this location is good because it is on the sidewalk, far from the person and the car, and not in the way"}, we would extract entities as \emph{`sidewalk'}, \emph{`person'}, and \emph{`car'}, and auxiliary concepts/predicates as \emph{`is\_on'}, \emph{`is\_far'}, and \emph{`in\_way'} and if any of the predicates is not present in the current $\conceptlibrary$, we'll first learn it (\ie applying \algoname\ recursively for hierarchical learning) by querying the user for a few demonstrations.

Algorithm \ref{alg:concept_lib_up} summarizes this discussion. In lines 3\--5, the \texttt{ExtractEntities} procedure uses an LLM to ground the entities used in the NL description and cross-reference them against existing entities in the concept library. Any new entities are added to the concept library without requiring user interaction, as we assume that any entity can be extracted from the observation using an open-set VLM. Lines 6\--17, on the other hand, extract new \emph{predicates} (auxiliary concepts) from the NL description and add it to the concept library. Since the semantics of these predicates cannot be assumed to be known a priori (unless they are already in the concept library), we must actively query the user to learn their semantics. Thus, the \texttt{QueryUserForDemonstration} procedure obtains new demonstrations, which are then used to synthesize these new predicates through recursive invocation of \texttt{Learn} at line 15, so that when the \texttt{UpdateConceptLibrary} procedure terminates, the new concept library $\conceptlibrary'$ contains all entities and predicates of interest.

\subsubsection{Program Sketch Synthesis.} Once \algoname\ has all the required concepts as part of its library, it uses an LLM to synthesize a \emph{program sketch}, which is a program with missing constants to be learned. We differentiate between program sketches and complete programs because the user's NL explanation is often sufficient to understand the general structure of the preference evaluation function but not its numeric parameters, which can only be accurately learned from the physical demonstrations. In particular, it first prompts the LLM to translate the NL explanation $e$ to a pair $(\Phi, r)$ where $\Phi$ is a formula in conjunctive normal form (CNF) over the predicates in the concept library and $r$ is the user's preference. Then, in a second step, \algoname\ prompts the LLM to update the previous sketch  $\hat{\pi}$ to a new one $\hat{\pi}'$ such that  $\hat{\pi}'$ returns $r$ when $\Phi$ evaluates to \texttt{True}. We found this two-stage process of first converting the NL explanation to a CNF formula and then prompting the LLM to repair the old sketch to work better in practice compared to prompting the LLM directly with all inputs (see Section~\ref{sec:evaluation}). For our running example, the $\Phi$ would be \emph{is\_on(`sidewalk') and is\_far(`person') and is\_far(`car') and not in\_way()}, and $r$ would be \emph{`good'}.

\subsubsection{Parameter Synthesis.} As mentioned earlier, a program sketch contains unknown numeric parameters that arise from the ambiguity of NL, \eg what does \emph{``close''} mean in terms of distances between objects? Thus, the last step of the \algoname\ pipeline utilizes the user's physical demonstrations to synthesize numeric parameters in the sketch. As summarized in Algorithm \ref{alg:param_synth}, it constructs a logical formula $\varphi$ consistent with all demonstrations as follows: first, for each physical demonstration $d$, it partially evaluates $\hat{\pi}$ (line 5) by fully evaluating all expressions without any unknowns, yielding a much simpler sketch containing only unknowns to be synthesized but no other variables. For instance, if the sketch contains the predicate \texttt{distanceTo(car)}, we can use the observation from $d$ to compute the actual distance between the location and the car. Next, let $\llbracket \hat{\pi} \rrbracket^i$ denote the condition under which $\hat{\pi}$ returns preference $i \in P$, and suppose that the current demonstration $d$ illustrates preference class $r$. Since we would like the synthesized program to return $r$ for demonstration $d$, $\llbracket \hat{\pi} \rrbracket^r$ should evaluate to \texttt{True}, while for all other preference classes $i$ where $i \neq r$, $\llbracket \hat{\pi} \rrbracket^i$ should evaluate to \texttt{False}. Thus, the loop in lines 8\--10 iteratively strengthens formula $\varphi$ by conjoining it with $\llbracket \hat{\pi} \rrbracket^r$ and the negation of $\llbracket \hat{\pi} \rrbracket^i$ for any $i$ distinct from $r$. Finally, we use an off-the-shelf constraint solver to obtain a model of the resulting formula\footnote{In general, the demonstrations may be noisy (\ie $\varphi$ could be unsatisfiable), which is quite often the case with real-world data. Thus, we use a MaxSMT solver~\cite{maxsmt} to maximize the number of satisfied clauses.} that is maximally consistent with the user's demonstrations. This results in a fully learned program that represents the user's preference.
\begin{figure*}[t]
\centering
\includegraphics[width=0.76\textwidth]{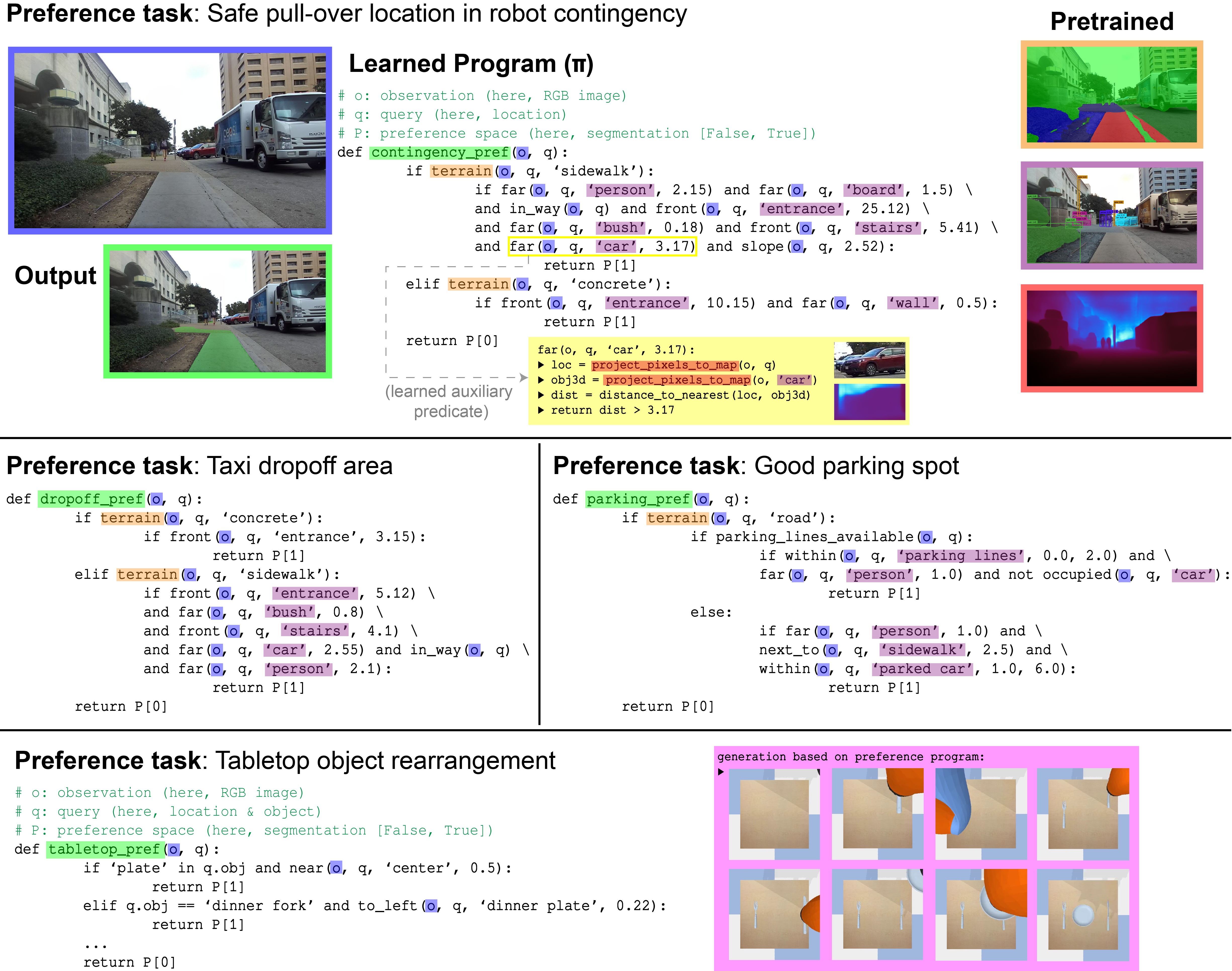}
\caption{\textbf{Preference tasks.} We show evaluation on three mobility tasks and one manipulation task. \algoname\ utilizes pretrained module outputs and executes the learned program.}
\label{fig:tasks}
\end{figure*}

\begin{figure}[tb]
  \centering
  \includegraphics[width=0.7\columnwidth]{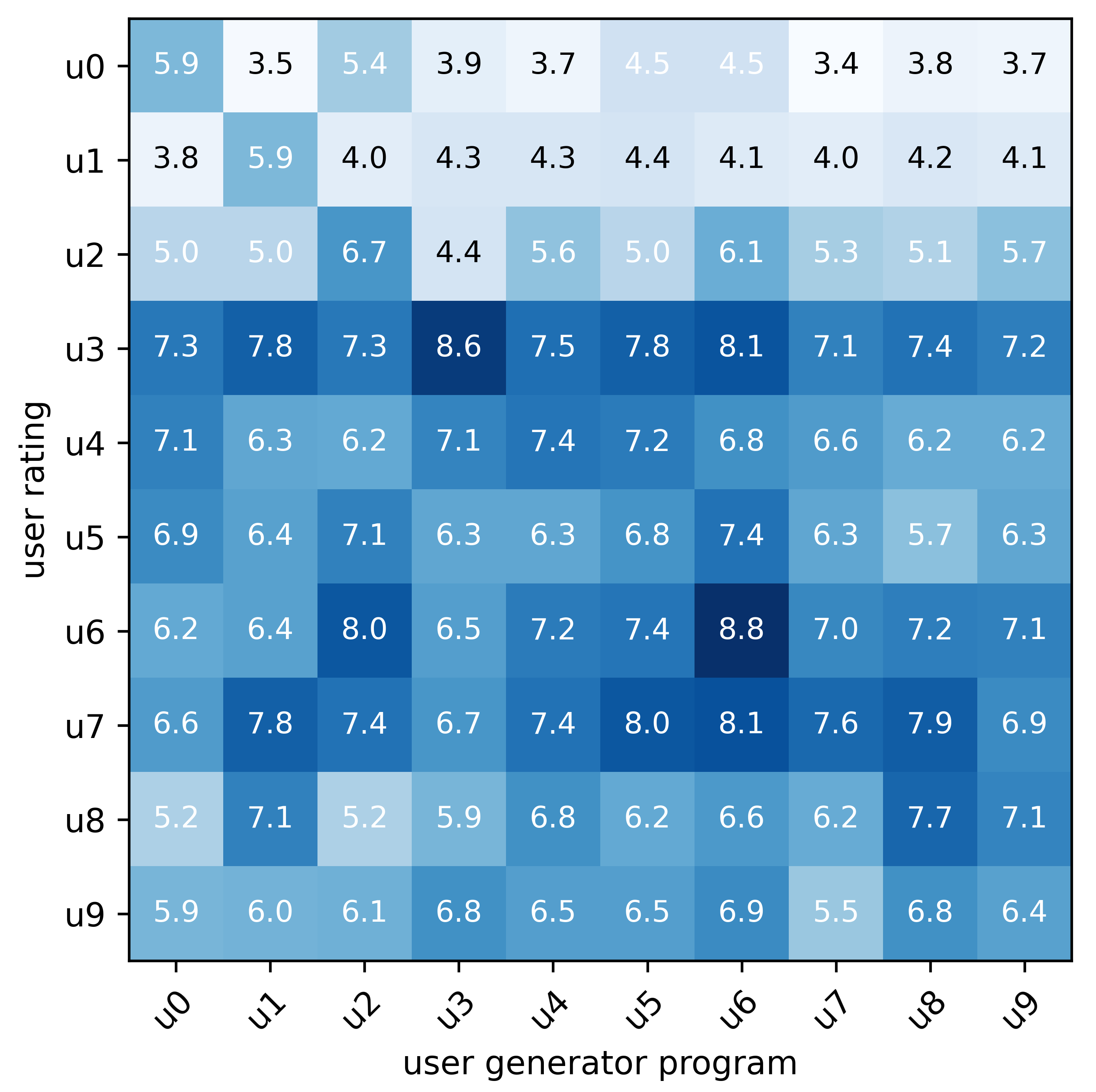}
  \caption{\textbf{User-study.} Higher entries around diagonal show good alignment between learned program and preference.}
  \label{fig:llmg_user}
\end{figure}

\section{Evaluation} \label{sec:evaluation}
We first describe the experimental setup and the benchmark for mobility tasks, and then present the performance of \algoname\ across four dimensions:
\begin{enumerate}
    \item[\textbf{Q1.}] How does its accuracy compare to other approaches?
    \item[\textbf{Q2.}] Can it easily and effectively extend to other domains?
    \item[\textbf{Q3.}] Can it align well to dissimilar multi-user preferences?
    \item[\textbf{Q4.}] How important are the various design choices?
\end{enumerate}
\subsubsection{Experimental Setup.} We evaluate on three mobility-related preferential concepts: a) \texttt{CONTINGENCY}: \emph{What is a good spot for a robot to pull over to in case of an emergency?}, b) \texttt{DROPOFF}: \emph{What is a good location for an autonomous taxi to stop and drop-off a customer?}, and c) \texttt{PARKING}: \emph{What is a good location for parking an autonomous car?}. In this work, we consider the preference space as binary only. The human demonstrations include the robot trajectories of the user driving the robot to the preferred location using a joystick, and NL description to explain the rationale for choosing that location. We use Grounded-SAM~\cite{GroundedSAM2024} zero-shot VLM for object detection, Depth Anything~\cite{yang2024depth} for zero-shot depth estimation, and a custom terrain model with SegFormer architecture~\cite{xie2021segformer} finetuned on custom data, since we observed that open-set visual models perform pretty poor on terrain segmentation. We use these models to get segmentation masks of the neural concepts (\ie objects and terrains) in the concept library. We use GPT-4~\cite{GPT42023} as the LLM for sketch synthesis. Lastly, as mentioned earlier, \algoname\ can interactively query the user to clarify new concepts that are present in the user's NL explanation but not in the current concept library. In principle, \algoname\ can query the user for both physical demonstrations and NL explanations. However, to reduce the burden on the user, \algoname, by default, only queries the user for NL explanations of auxiliary concepts and performs synthesis of auxiliary concepts using NL explanations alone.

\subsubsection{Baselines.} We create a dataset of 815 labeled images taken from the \anon{UT Austin campus} area, where the labels mark the locations on the images that are consistent with the intended user preference for each of the mobility tasks. We split the dataset into three sets: train, in-distribution test, and out-of-distribution test sets. The train and in-distribution sets belong to the same geographical region, while the out-of-distribution set belongs to a different region. Table \ref{tab:iou_baselines} shows the comparison against various baselines (we only show the strongest variant of each baseline here). We use mean Intersection-Over-Union (mIOU) as the metric and evaluate the following baselines: (1) pure neural models based on SegFormer (SF)~\cite{xie2021segformer} architecture (with and without depth input) and DinoV2~\cite{dinov2}, both with pretrained weights and fine-tuned on our custom dataset; (2) NS-CL~\cite{NSCL2019}, a neurosymbolic concept learning approach for predominantly \emph{factual} concepts, trained on our dataset; (3) \texttt{VisProg}~\cite{VisProg2023} which is a state-of-the-art VQA neurosymbolic method; and (4) \texttt{GPT4}~\cite{GPT42023} vision. We find that \algoname\ outperforms all baselines and improves on the closest baseline by a significant margin on out-of-distribution test data -- 74.07 \vs{} 57.42 for \texttt{CONTINGENCY}, 80.72 \vs{} 63.99 for \texttt{DROPOFF}, and 62.76 \vs{} 52.91 for \texttt{PARKING}. Further, even though \algoname\ is trained on an order of magnitude fewer samples (for instance, 29 demonstrations for \texttt{CONTINGENCY}) than neural baselines (for instance, 224 images for \texttt{CONTINGENCY}), it matches or improves the baseline.


\newcommand{\cell}[2]{#1}
\begin{table*}[t]
    \centering
  \begin{tabular}{l*{9}{c}}
        \toprule
        & \multicolumn{3}{c}{\textsc{CONTINGENCY}} & \multicolumn{3}{c}{\textsc{DROPOFF}} & \multicolumn{3}{c}{\textsc{PARKING}} \\
        \cmidrule(lr){2-4} \cmidrule(lr){5-7} \cmidrule(l){8-10}
        Method / Split & train & in-test & out-test & train & in-test & out-test & train & in-test & out-test \\
        \midrule
        SYNAPSE & \cell{\bfseries 77.64}{1.41} & \cell{\bfseries 76.29}{1.06} & \cell{\bfseries 74.07}{1.89} & \cell{\bfseries 79.32}{1.21} & \cell{\bfseries 80.18}{1.45} & \cell{\bfseries 80.72}{1.34} & \cell{68.60}{1.12} & \cell{\bfseries 66.87}{1.23} & \cell{\bfseries 62.76}{0.85} \\
        \midrule
        SF-RGBD-b5 & \cell{76.48}{0.00} & \cell{67.81}{0.00} & \cell{56.11}{0.00} & \cell{77.69}{0.00} & \cell{70.70}{0.00} & \cell{52.39}{0.00} & \cell{\bfseries 71.06}{0.00} & \cell{65.72}{0.00} & \cell{49.99}{0.00} \\
        DinoV2-g & \cell{73.65}{0.00} & \cell{60.93}{0.00} & \cell{51.23}{0.00} & \cell{79.50}{0.00} & \cell{72.17}{0.00} & \cell{59.10}{0.00} & \cell{67.06}{0.00} & \cell{62.04}{0.00} & \cell{52.78}{0.00} \\
        \midrule
        NS-CL & \cell{69.76}{0.00} & \cell{69.63}{0.00} & \cell{63.65}{0.00} & \cell{71.26}{0.00} & \cell{70.38}{0.00} & \cell{63.99}{0.00} & \cell{46.92}{0.00} & \cell{43.71}{0.00} & \cell{45.23}{0.00} \\
        VisProg & \cell{38.94}{3.64} & \cell{39.21}{2.64} & \cell{41.83}{2.11} & \cell{39.17}{2.95} & \cell{39.44}{2.67} & \cell{43.14}{3.04} & \cell{38.88}{3.40} & \cell{39.64}{2.80} & \cell{38.99}{2.95} \\
        GPT4V & \cell{28.73}{3.23} & \cell{28.96}{4.34} & \cell{33.92}{4.98} & \cell{39.38}{4.27} & \cell{38.34}{4.62} & \cell{39.14}{3.18} & \cell{41.38}{3.56} & \cell{42.20}{4.01} & \cell{39.77}{4.09} \\
        \bottomrule
    \end{tabular}
    \caption{Mean IOU (\%) $\uparrow$ results for the three concepts. The train set represents the full set -- \algoname\ only needs 29 demonstrations (from the train area), while other fine-tuned (SegFormer, DinoV2) or trained (NS-CL) baselines use the full set.}
  \label{tab:iou_baselines}
\end{table*}

\begin{table*}[t]
    \centering
    \setlength\tabcolsep{2pt}
  \begin{tabular}{l*{6}{c}}
        \toprule
        Method / Feature & feat1 & feat2 & \textsc{LLM} & \textsc{VLM} & \texttt{mIOU (\%)} \\
        \midrule
        SYNAPSE        & \tick & \tick & GPT-4~\citep{GPT42023} & DINO-SAM~\citep{GroundedSAM2024} & {\bfseries 76.11} \\
        \midrule
        SYNAPSE-SynthDirect     & \cros & \cros & GPT-4~\citep{GPT42023} & DINO-SAM~\citep{GroundedSAM2024} & 60.74 \\
        SYNAPSE-SynthCaP       & \cros & \tick & GPT-4~\citep{GPT42023} & DINO-SAM~\citep{GroundedSAM2024} & 64.11 \\
        \midrule
        SYNAPSE-PaLM2         & \tick & \tick & PaLM2~\citep{palm2} & DINO-SAM~\citep{GroundedSAM2024} & 71.62 \\
        SYNAPSE-GroupViT     & \tick & \tick & GPT-4~\citep{GPT42023} & GroupViT~\citep{xu2022groupvit} & 73.41 \\
        \midrule
        SF-RGBD-b5 & - & - & - & - & 53.81 \\
        DinoV2-g & - & - & - & - & 65.71 \\
        \bottomrule
    \end{tabular}
    \caption{The results for the ablation studies. Evaluation is on the full \texttt{CONTINGENCY} dataset.}
  \label{tab:ablations_features}
\end{table*}


\begin{table}[t]
    \centering
    \setlength{\tabcolsep}{1mm}
  \begin{tabular}{l*{8}{c}}
        \toprule
        Task \#ID & 1 & 2 & 3 & 4 & 5 & 6 & 7 & 8 \\
        \midrule
        SYNAPSE & \bfseries 7.13 & \bfseries 6.57 & \bfseries 5.67 & \bfseries 7.63 & \bfseries 7.23 & \bfseries 6.73 & \bfseries 8.30 & \bfseries 6.50\\
        \midrule
        LLM-GROP   & 4.07 & 3.27 & 3.37 & 5.83 & 4.40 & 4.50 & 5.80 & 2.70\\
        LATP       & 3.93 & 1.70 & 2.60 & 2.10 & 3.23 & 2.93 & 1.93 & 2.33\\
        GROP       & 2.60 & 2.47 & 2.87 & 2.37 & 2.37 & 2.77 & 3.57 & 2.33\\
        TPRA       & 2.77 & 2.27 & 2.47 & 2.17 & 2.87 & 2.17 & 2.13 & 2.07\\
        \bottomrule
    \end{tabular}
    \caption{User ratings on LLM-GROP~\cite{LLMGROP2023} tabletop object rearrangement task on a scale of 1-10.}
    \label{tab:ratings_llmgrop}
\end{table}

\subsubsection{Generalization to other domains.} We evaluate \algoname\ on the tabletop object arrangement task: \emph{Given a set of objects on dinner table, what is a good arrangement?}, as introduced in LLM-GROP~\cite{LLMGROP2023}, to show extension to other domains such as robot manipulation. Similar to the LLM-GROP work, we use 10 participants and utilise user ratings as the metric. This task consists of 8 sub-tasks with different sets of objects. We use five of these as our train tasks and the rest three as the test tasks. We collect one demonstration per train task from each user, where again the demonstration consists of the user showing the preferred object arrangement as well as a NL description. We use the baselines from LLM-GROP. Table~\ref{tab:ratings_llmgrop} summarizes the results where \algoname\ outperforms the closest baseline by an average of 2.7 points across all tasks.

\subsubsection{Multi-user preferences.} For the LLM-GROP task, since we have learned preference programs for all 10 participants, we test the alignment of the learned programs with the different user preferences. For this, we generate object arrangements using learned programs for each user and then ask all other users to rate the arrangement. The results are summarized in Figure~\ref{fig:llmg_user} which shows the average rating across the eight sub-tasks. For each user, the highest performance is attained by the program that was learned from the same user's demonstrations, which indicates good alignment.

\subsubsection{Ablations.} We investigate four classes of ablations: (1) \emph{NN-ablations}, in which we compare the performance of neural baselines (SF and DinoV2) against \algoname\ when trained on the \emph{same} number of samples (\ie 29); (2) \emph{LLM-based} in which we replace GPT-4 with different models in program synthesis part of the framework; (3) \emph{VLM-based} ablations, where we test different VLMs for object detection in our framework; and (4) \emph{framework} ablations where we test the following framework features: (a) \emph{feat1:} whether it queries the user for auxiliary concepts, (b) \emph{feat2:} whether it performs lifelong learning by building on its concept library. It can be seen from Table~\ref{tab:ablations_features} that the NN-ablations perform poorly since they are exposed to so few training samples that they aren't able to generalize well to the full dataset. Changing the program synthesis process of \algoname\ (\ie not maintaining the library) or the LLM/VLM which in turn affects the accuracy of the program sketch and the parameters being synthesized, respectively, also has a significant impact on the performance.

More details on the evaluation and additional experiments are provided in the supplementary material.
\section{Conclusion, Limitations \& Future work} \label{sec:discussion}
We presented \algoname, a data-efficient, neuro-symbolic framework for learning preferential concepts from a small number of human demonstrations. We experimentally showed that \algoname\ achieves strong generalization on new data and it outperforms the baselines by a large margin ($\approx 15\%$ mIOU). Further, we showed that \algoname\ is able to align well with multi-user preferences. Finally, we also showed that \algoname\ extends to other domains effectively. However, \algoname\ has some potential limitations as well. First, \algoname\ relies substantially on the quality of underlying neural modules and their capabilities. In our experiments, we observe that a careful selection of parameters and clever prompting is needed to achieve best performance. Further, \algoname\ relies on the quality of the user's NL utterance as well as physical demonstrations for accurate synthesis. In practice, the demonstrations could be noisy and imperfect. Although \algoname\ tries to compensate for slight inconsistencies in the user demonstration by using MaxSMT, however, to truly tackle this noise, a probabilistic approach to neurosymbolic programs needs to be explored. Finally, \algoname\ as presented here doesn't take into account dynamically varying preferences, \ie if a person's preference changes drastically between 1st and the nth sample, \algoname\ would still give equal weight to both. A recency weighting approach might resolve this limitation.



\section*{Acknowledgements}
This work is partially supported by the National Science Foundation (CAREER-2046955, OIA-2219236, DGE-2125858, CCF-2319471, CCF-1918889,  CCF-1901376). Any opinions, findings, and conclusions expressed in this material are those of the authors and do not necessarily reflect the views of the sponsors.

\bibliography{aaai25}
\clearpage
\appendix
\begin{figure}
\centering
\includegraphics[width=0.95\linewidth]{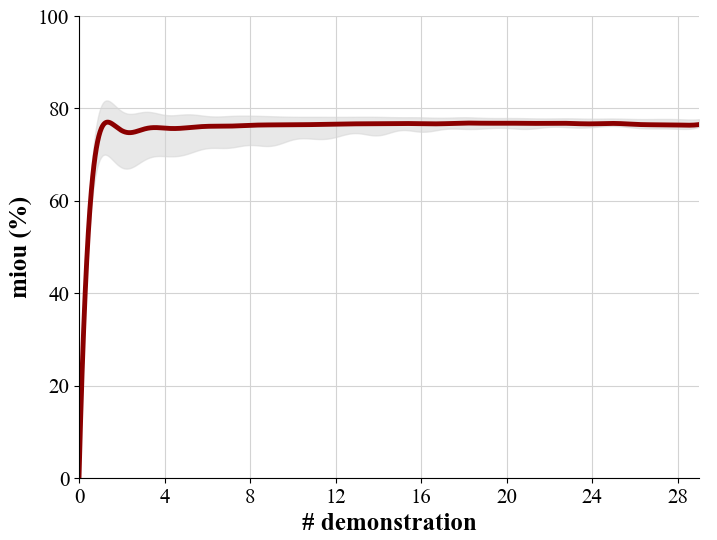}
\caption{Plot showing susceptibility of \algoname\ to reordering of demonstrations. \emph{Gray} area represents the mean IOU (\%) variation as \algoname\ sees more demonstrations.}
\label{fig:appendix_reorder}
\end{figure}

\begin{figure}
\centering
\includegraphics[width=0.75\columnwidth]{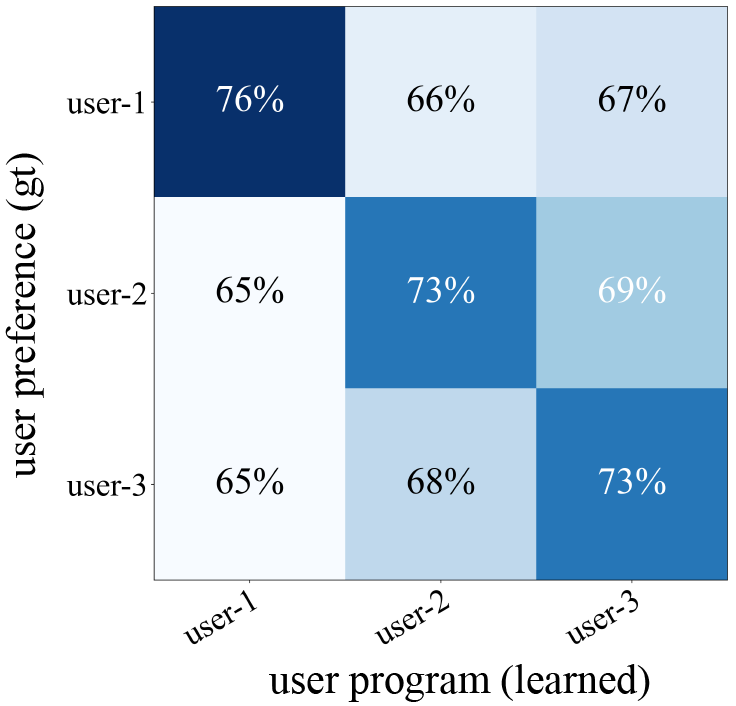}
\caption{\textbf{User study results for \texttt{CONTINGENCY}.} We run the learned programs for each user on the preference dataset of each other user and report mIOU. Higher entries along the diagonal indicates good alignment of the learned program with the corresponding user preference.}
\label{fig:userstudy}
\end{figure}

\begin{figure*}[t]
\centering
\includegraphics[width=0.9\textwidth]{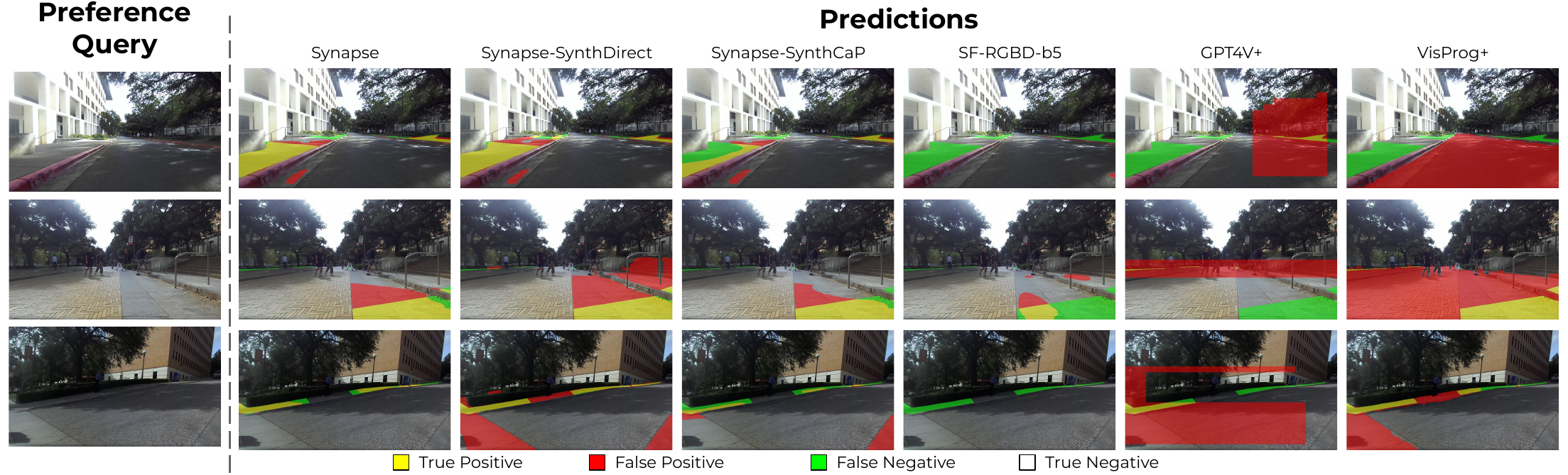}
\caption{An illustrative comparison between \algoname, the baselines, and ablations for \texttt{CONTINGENCY}. Color coding shows the overlap of the predictions with the \emph{ground-truth}.}
\label{fig:appendix_baselines_exs}
\end{figure*}

\begin{figure*}[t]
\centering
\includegraphics[width=0.95\textwidth]{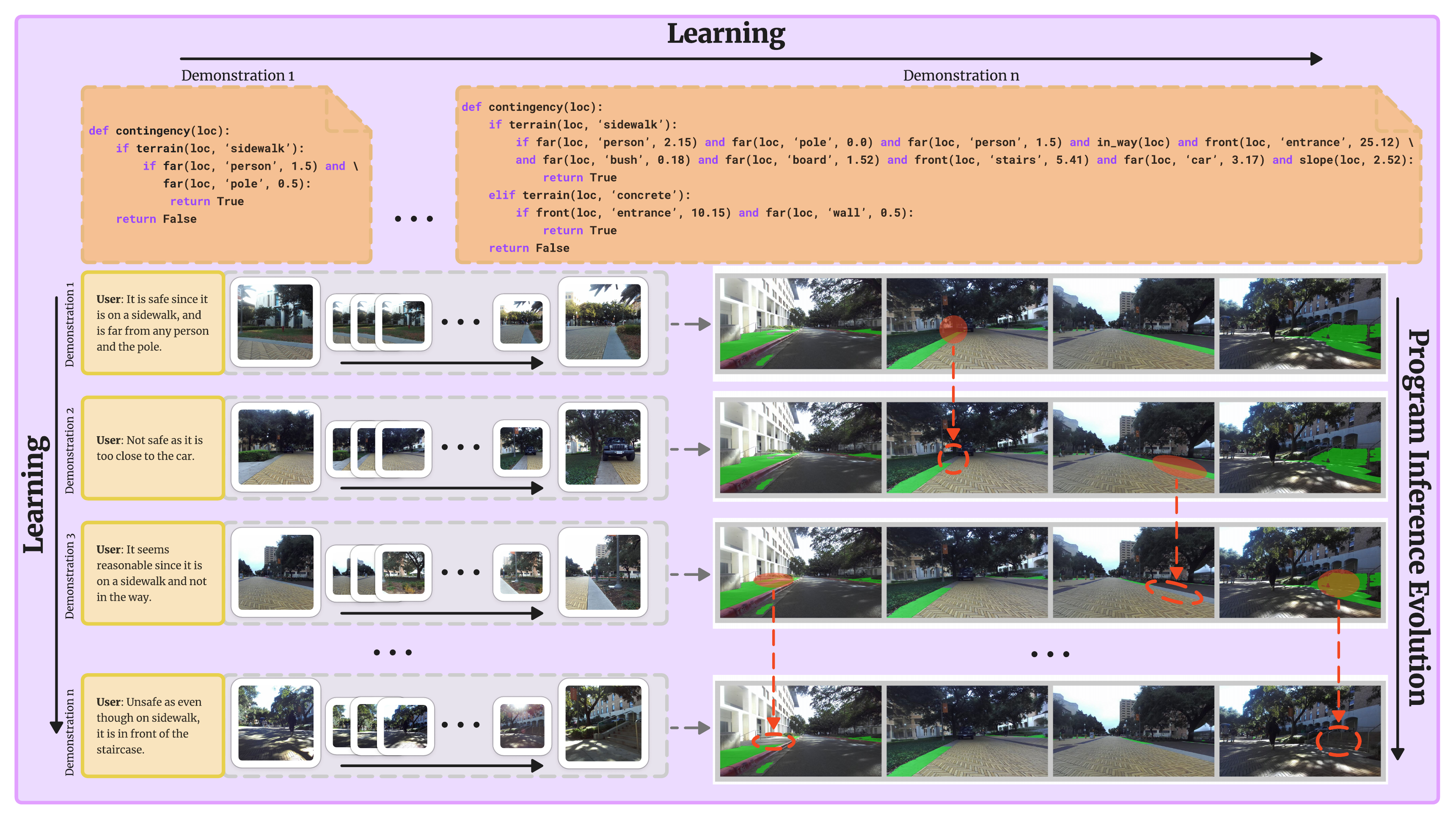}
\caption{Overview of \algoname\ learning and program inference evolution. It shows the \emph{lifelong} learning characteristic.}
\label{fig:appendix_lifelong}
\end{figure*}

\begin{figure*}[t]
\centering
\includegraphics[width=0.95\textwidth]{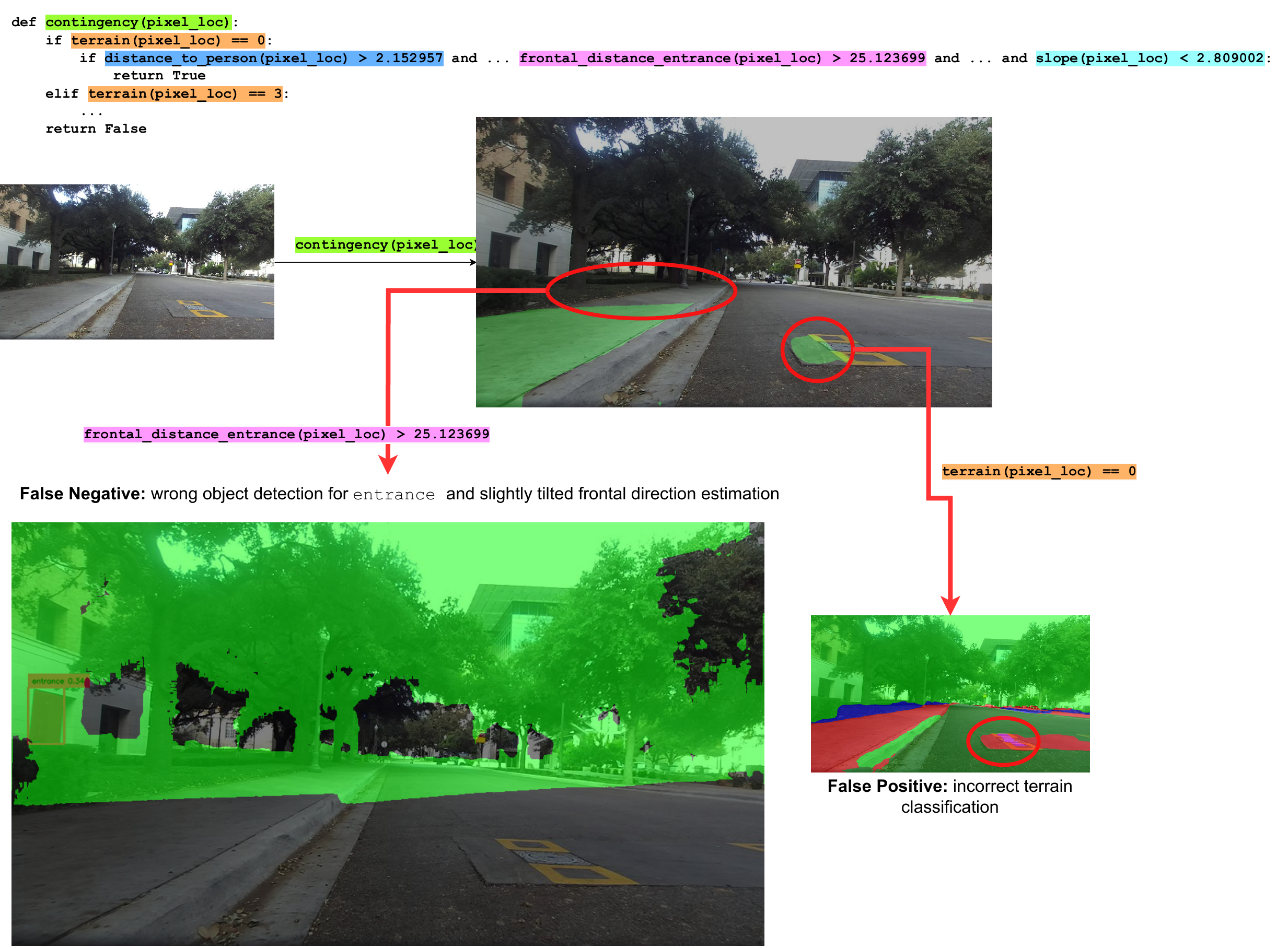}
\caption{An illustration of reasoning about failures using the learned program via backtracking.}
\label{fig:appendix_error_analy}
\end{figure*}

\begin{figure*}[t]
\centering
\includegraphics[width=0.95\textwidth]{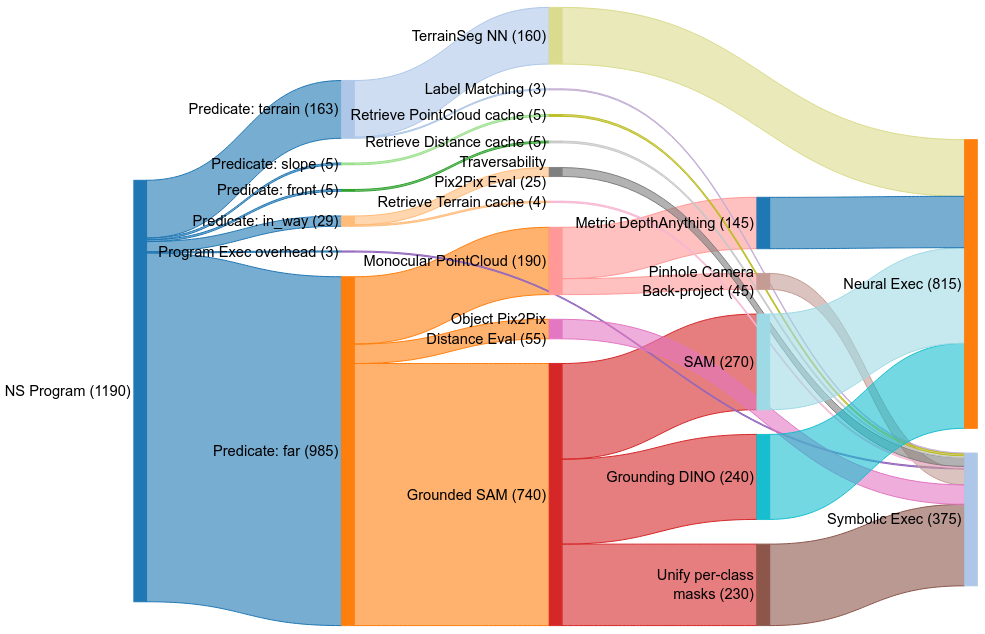}
\caption{Average real-world deployment runtime (milliseconds) split for a single GPU single process execution for \texttt{CONTINGENCY}. It runs at about 1 Hz.}
\label{fig:appendix_sankey}
\end{figure*}

\begin{table}[t]
\centering
\setlength{\tabcolsep}{1mm}
\begin{tabular}{lccc}
    \toprule
        & box-thresh. & text-thresh. & nms-thresh. \\
        \midrule
        barricade          & 0.5       & 0.5       & 0.3 \\
        board              & 0.3       & 0.3       & 0.5 \\
        bush               & 0.4       & 0.4       & 0.4 \\
        car                & 0.3       & 0.3       & 0.3 \\
        entrance           & 0.3       & 0.3       & 0.2 \\
        person             & 0.25       & 0.25       & 0.6 \\
        pole               & 0.4       & 0.4       & 0.5 \\
        staircase          & 0.25       & 0.25       & 0.4 \\
        tree               & 0.4       & 0.4       & 0.45 \\
        wall               & 0.5       & 0.5       & 0.4 \\
        \bottomrule
\end{tabular}
\caption{\textbf{Grounded-SAM~\cite{GroundedSAM2024}.} Choosing common parameters $(0.3, 0.3, 0.4)$ irrespective of the object category works fine, though we observed to achieve best performance on the version at the time, these object-specific parameters were needed. For any new object class, we use the \emph{closest} category's parameters.}
\label{tab:appendix_sam_hparams}
\end{table}

\begin{table}[t]
\centering
\begin{tabular}{lc}
    \toprule
        key & value \\
        \midrule
        seed & 0 \\
        temperature & 0.0 \\
        stop & `END' \\
        \bottomrule
\end{tabular}
\caption{\textbf{LLMs.} Common hyperparameters for all language models.}
\label{tab:appendix_llm_hparams}
\end{table}

\begin{table*}[t]
    \centering
    \setlength\tabcolsep{2pt}
  \begin{tabular}{l*{6}{c}}
        \toprule
        Method / Feature & feat1 & feat2 & feat3 & \textsc{LLM} & \textsc{VLM} & \texttt{mIOU (\%)} \\
        \midrule
        Synapse        & \tick & \tick & \cros & GPT-4~\citep{GPT42023} & DINO-SAM~\citep{GroundedSAM2024} & {\bfseries 76.11} \\
        \midrule
        Synapse-SynthDirect     & \cros & \cros & \cros & GPT-4~\citep{GPT42023} & DINO-SAM~\citep{GroundedSAM2024} & 60.74 \\
        Synapse-SynthDirect+    & \cros & \cros & \tick & GPT-4~\citep{GPT42023} & DINO-SAM~\citep{GroundedSAM2024} & 68.86 \\
        Synapse-SynthCaP       & \cros & \tick & \cros & GPT-4~\citep{GPT42023} & DINO-SAM~\citep{GroundedSAM2024} & 64.11 \\
        \midrule
        Synapse-CodeLLama     & \tick & \tick & \cros & CodeLLama~\citep{codellama} & DINO-SAM~\citep{GroundedSAM2024} & 69.88 \\
        Synapse-StarCoder     & \tick & \tick & \cros & StarCoder~\citep{starcoder} & DINO-SAM~\citep{GroundedSAM2024} & 63.62 \\
        Synapse-PaLM2         & \tick & \tick & \cros & PaLM2~\citep{palm2} & DINO-SAM~\citep{GroundedSAM2024} & 71.62 \\
        \midrule
        Synapse-OWLViTSAM     & \tick & \tick & \cros & GPT-4~\citep{GPT42023} & OWLViT~\citep{owlvit} & 70.17 \\
        Synapse-GroupViT     & \tick & \tick & \cros & GPT-4~\citep{GPT42023} & GroupViT~\citep{xu2022groupvit} & 73.41 \\
        \midrule
        SF-RGB-b0 & - & - & - & - & - & 44.58 \\
        SF-RGB-b5 & - & - & - & - & - & 46.30 \\
        SF-RGBD-b0 & - & - & - & - & - & 45.84 \\
        SF-RGBD-b5 & - & - & - & - & - & 53.81 \\
        DinoV2-b & - & - & - & - & - & 57.05 \\
        DinoV2-g & - & - & - & - & - & 65.71 \\
        \bottomrule
    \end{tabular}
    \caption{Full ablation study results on \texttt{CONTINGENCY}. We report the mean across five runs.}
  \label{tab:appendix_ablations}
\end{table*}

\begin{table*}[t]
    \centering
  \begin{tabular}{l*{9}{c}}
        \toprule        
        & \multicolumn{3}{c}{iou\_pos (\%)} & \multicolumn{3}{c}{iou\_neg (\%)} & \multicolumn{3}{c}{miou (\%)} \\
        \cmidrule(lr){2-4} \cmidrule(lr){5-7} \cmidrule(l){8-10}
        & train & in-test & out-test & train & in-test & out-test & train & in-test & out-test \\
        \midrule
        Synapse & 57.22 & \bfseries 54.46 & \bfseries 50.18 & \bfseries 98.06 & \bfseries 98.11 & \bfseries 97.96 & \bfseries 77.64 & \bfseries 76.29 & \bfseries 74.07 \\
        \midrule
        DinoV2-b & 57.86 & 42.69 & 20.33 & 77.33 & 69.24 & 66.87 & 67.60 & 55.96 & 43.60 \\
        DinoV2-g & \bfseries 61.80 & 46.43 & 24.04 & 85.50 & 75.42 & 78.42 & 73.65 & 60.93 & 51.23 \\
        SF-RGB-b0 & 43.54 & 28.46 & 17.77 & 97.44 & 97.03 & 97.07 & 70.49 & 62.75 & 57.42 \\
        SF-RGB-b5 & 51.63 & 43.87 & 19.04 & 97.54 & 97.08 & 92.96 & 74.59 & 70.48 & 56.00 \\
        SF-RGBD-b0 & 46.71 & 37.07 & 13.49 & 97.62 & 97.39 & 96.00 & 72.17 & 67.23 & 54.75 \\
        SF-RGBD-b5 & 55.10 & 38.48 & 15.71 & 97.85 & 97.13 & 96.50 & 76.48 & 67.81 & 56.11 \\
        \midrule
        NS-CL & 42.21 & 41.87 & 30.18 & 97.31 & 97.38 & 97.12 & 69.76 & 69.63 & 63.65 \\
        GPT4V & 01.73 & 01.91 & 02.74 & 51.38 & 57.51 & 58.15 & 26.56 & 29.71 & 30.45 \\
        GPT4V+ & 02.29 & 02.18 & 02.59 & 55.16 & 55.74 & 65.24 & 28.73 & 28.96 & 33.92 \\
        VisProg & 04.99 & 01.25 & 05.04 & 86.24 & 90.01 & 84.92 & 45.62 & 45.63 & 44.98 \\
        VisProg+ & 08.13 & 06.36 & 07.15 & 69.75 & 72.06 & 76.51 & 38.94 & 39.21 & 41.83 \\
        \bottomrule
    \end{tabular}
    \caption{Full mean IOU (\%) $\uparrow$ results for \texttt{CONTINGENCY}. We report the mean across five runs.}
  \label{tab:appendix_contingency}
\end{table*}

\begin{table*}[t]
    \centering
  \begin{tabular}{l*{9}{c}}
        \toprule        
        & \multicolumn{3}{c}{iou\_pos (\%)} & \multicolumn{3}{c}{iou\_neg (\%)} & \multicolumn{3}{c}{miou (\%)} \\
        \cmidrule(lr){2-4} \cmidrule(lr){5-7} \cmidrule(l){8-10}
        & train & in-test & out-test & train & in-test & out-test & train & in-test & out-test \\
        \midrule
        Synapse & \bfseries 60.64 & \bfseries 62.05 & \bfseries 63.13 & 97.99 & \bfseries 98.31 & \bfseries 98.31 & 79.32 & \bfseries 80.18 & \bfseries 80.72 \\
        \midrule
        DinoV2-b & 56.53 & 42.83 & 31.87 & 79.72 & 69.23 & 58.15 & 68.12 & 56.03 & 45.01 \\
        DinoV2-g & 60.31 & 46.59 & 34.28 & \bfseries 98.69 & 97.74 & 83.92 & \bfseries 79.50 & 72.17 & 59.10 \\
        SF-RGB-b0 & 48.59 & 38.93 & 08.12 & 97.39 & 97.32 & 96.30 & 72.99 & 68.13 & 52.21 \\
        SF-RGB-b5 & 56.73 & 48.00 & 15.94 & 97.78 & 97.66 & 94.13 & 77.26 & 72.83 & 55.04 \\
        SF-RGBD-b0 & 51.09 & 42.17 & 13.89 & 97.57 & 97.43 & 95.90 & 74.33 & 69.80 & 54.90 \\
        SF-RGBD-b5 & 57.43 & 43.86 & 08.93 & 97.95 & 97.54 & 95.84 & 77.69 & 70.70 & 52.39 \\
        \midrule
        NS-CL & 45.16 & 43.42 & 31.01 & 97.36 & 97.33 & 96.97 & 71.26 & 70.38 & 63.99 \\
        GPT4V & 02.40 & 02.25 & 03.09 & 58.67 & 63.69 & 72.20 & 30.54 & 32.97 & 37.65 \\
        GPT4V+ & 04.43 & 01.90 & 03.02 & 74.33 & 74.77 & 75.25 & 39.38 & 38.34 & 39.14 \\
        VisProg & 01.48 & 01.94 & 06.58 & 91.10 & 93.03 & 89.56 & 46.29 & 47.49 & 48.07 \\
        VisProg+ & 08.62 & 06.86 & 08.70 & 69.71 & 72.01 & 77.57 & 39.17 & 39.44 & 43.14 \\
        \bottomrule
    \end{tabular}
    \caption{Full mean IOU (\%) $\uparrow$ results for \texttt{DROPOFF}. We report the mean across five runs.}
  \label{tab:appendix_dropoff}
\end{table*}

\begin{table*}[t]
    \centering
  \begin{tabular}{l*{9}{c}}
        \toprule
        & \multicolumn{3}{c}{iou\_pos (\%)} & \multicolumn{3}{c}{iou\_neg (\%)} & \multicolumn{3}{c}{miou (\%)} \\
        \cmidrule(lr){2-4} \cmidrule(lr){5-7} \cmidrule(l){8-10}
        & train & in-test & out-test & train & in-test & out-test & train & in-test & out-test \\
        \midrule
        Synapse & 39.14 & \bfseries 34.87 & \bfseries 27.24 & 98.06 & 98.87 & \bfseries 98.27 & 68.60 & \bfseries 66.87 & \bfseries 62.76 \\
        \midrule
        DinoV2-b & 32.65 & 29.12 & 20.89 & 95.97 & 98.12 & 93.75 & 64.31 & 63.62 & 57.32 \\
        DinoV2-g & 37.21 & 30.10 & 22.36 & 96.91 & 93.98 & 83.20 & 67.06 & 62.04 & 52.78 \\
        SF-RGB-b0 & 21.28 & 15.28 & 04.91 & 94.07 & 98.50 & 94.40 & 57.68 & 56.89 & 49.66 \\
        SF-RGB-b5 & 38.99 & 24.86 & 08.13 & \bfseries 98.27 & 99.42 & 97.68 & \bfseries 68.63 & 62.14 & 52.91 \\
        SF-RGBD-b0 & 24.80 & 17.49 & 04.67 & 97.00 & 96.95 & 96.70 & 60.90 & 57.22 & 50.69 \\
        SF-RGBD-b5 & \bfseries 43.97 & 31.88 & 03.50 & 98.14 & \bfseries 99.56 & 96.48 & 71.06 & 65.72 & 49.99 \\
        \midrule
        NS-CL & 11.34 & 05.54 & 05.28 & 82.51 & 81.88 & 85.18 & 46.92 & 43.71 & 45.23 \\
        GPT4V & 01.10 & 01.23 & 01.11 & 75.72 & 75.01 & 79.25 & 38.41 & 38.12 & 40.18 \\
        GPT4V+ & 01.56 & 01.01 & 00.64 & 81.19 & 83.39 & 78.90 & 41.38 & 42.20 & 39.77 \\
        VisProg & 04.72 & 03.72 & 01.69 & 74.13 & 72.5 & 80.29 & 39.43 & 38.11 & 40.99 \\
        VisProg+ & 03.31 & 04.22 & 03.08 & 74.44 & 75.06 & 74.89 & 38.88 & 39.64 & 38.99 \\
        \bottomrule
    \end{tabular}
    \caption{Full mean IOU (\%) $\uparrow$ results for \texttt{PARKING}. We report the mean across five runs.}
  \label{tab:appendix_parking}
\end{table*}

\section{Implementation}
\subsubsection{Inputs.} For learning the programs, we collected a minimum of 10 demonstrations (upto 30) where each demonstration had a trajectory of robot poses and the associated RGB camera image. A few of the natural language prompts from the user are depicted in figure \ref{fig:appendix_lifelong}. However, for the LLM-GROP task, we only collect one demonstration for each of the 5 train sub-tasks, where each of these demonstrations has the preferred tabletop arrangement and the associated NL explanation.

\subsubsection{Pretrained models.} We use the following in our implementation:
\begin{enumerate}[label=\alph*.]
    \item Grounded-SAM~\cite{GroundedSAM2024}: We use Grounded-SAM with the hyperparameters as shown in table \ref{tab:appendix_sam_hparams} to perform zero-shot object detection and segmentation on images.
    \item Terrain segmentation: Our experiments showed that present VLMs do not do so well on terrain segmentation, which was a domain-specific essential capability to be able to represent the preferential concept well enough. Thus, we finetuned the SegFormer-b5 \cite{xie2021segformer} model, with pretrained weights from the HuggingFace transformers library, on our custom dataset.
    \item We use GPT-4~\cite{GPT42023} as our language model for the sketch synthesis module. Hyperparameters are shown in table \ref{tab:appendix_llm_hparams}. Prompts for doing different tasks in our framework (\ie, grounding, synthesis \etc) can be found in the codebase. Note that they have certain placeholders like $<!\ldots!>$ and $<dyn!\ldots!dyn>$ which refer to other prompt instances or are replaced dynamically in the code based on generated outputs.
    \item We use Depth Anything~\cite{yang2024depth} model to get zero-shot depth estimation from RGB images. This is required for 2D to 3D mapping, as well as for training depth-NN baselines.
\end{enumerate}

\section{Evaluation}
\subsection{Baselines}
We evaluate the following baselines:
\begin{enumerate}[label=*]
    \item SegFormer b0 and b5 models, both with or without depth input. For taking in depth, we only modify the input layer, and still retain all other pretrained weights. We take measures such as early stopping to prevent overfitting.
    \item DinoV2~\cite{dinov2} base and large variation models are finetuned on our custom dataset.
    \item NS-CL~\cite{NSCL2019}, a related approach in neuro-symbolic concept learning, is adapted for our task. For each individual predicate of NS-CL, we initialize it with DinoV2-b weights and then finetune it on our dataset.
    \item VisProg and VisProg+ come from a related approach to VQA \cite{VisProg2023}. We again follow the same methodology of prompting as for GPT4-vision.
    \item GPT4-vision: Due to token limitations, we query GPT4-vision to output a $20 \times 20$ output class array given the image. For reporting the IOU for GPT4-vision, we downsample the ground-truth to the same size for the comparison to be fair. The `+' variant essentially means that we prompt it with additional information about the user's ground-truth, \ie, we provide it the program that \algoname\ has learned, in natural language.
\end{enumerate}
\subsection{Ablations}
We test the following ablations for our framework:
\begin{enumerate}[label=*]
    \item Framework ablations: We test three alternative ways to generate the program sketch from given natural language input from the user:
    \begin{enumerate}[label=\alph*.]
        \item \texttt{Synapse-SynthDirect}: given only the basic predicates, we adopt a one-step approach to synthesis, where we ask LLM to update the program sketch based on the new NL input and previous program sketch, while utilising only the basic predicates, \ie, it has no way to hierarchically build and retain higher-level predicates, as well as it does not do any CNF extraction.
        \item \texttt{Synapse-SynthDirect+}: we provide the higher-level concepts learned by our main framework and then given these already learned higher-level predicates, we again adopt a one-step approach to synthesis, where we ask LLM to update the program sketch based on the new NL input and previous program sketch, \ie, it still does not do any CNF extraction. Note, however, this is \emph{only} possible for a post-learning ablation study, as in an actual learning framework, we do not have apriori access to the higher-level learned predicates.
        \item \texttt{Synapse-SynthCaP}: we disallow the framework from querying the user for auxiliary demonstrations to learn auxiliary concepts, \ie, the framework is forced to generate code (using the concept library) for the auxiliary concepts solely based on the information available, which essentially is the \emph{name} of that particular concept. This is similar to the recursion performed in Code-as-Policies \cite{CodeAsPolicies2023}.
    \end{enumerate}
    \item NN-ablations: We finetune the four SegFormer models and two DinoV2 models on the same number of samples as \algoname, which is 29 for the \texttt{CONTINGENCY} concept.
    \item LLM and VLM ablations: We test the performance of our overall framework using some of the other language models: (1) CodeLLama~\cite{codellama}, (2) StarCoder~\cite{starcoder}, and (3) PaLM2~\cite{palm2}, and a few vision-language models: (1) OWL-ViT~\cite{owlvit} with SAM~\cite{SAM2023}, and (2) GroupViT~\cite{xu2022groupvit}. We see that the performance of \algoname\ depends a fair bit on the strength of the underlying foundation models.
\end{enumerate}

\begin{figure*}[tb]
  \centering
  \includegraphics[width=\textwidth]{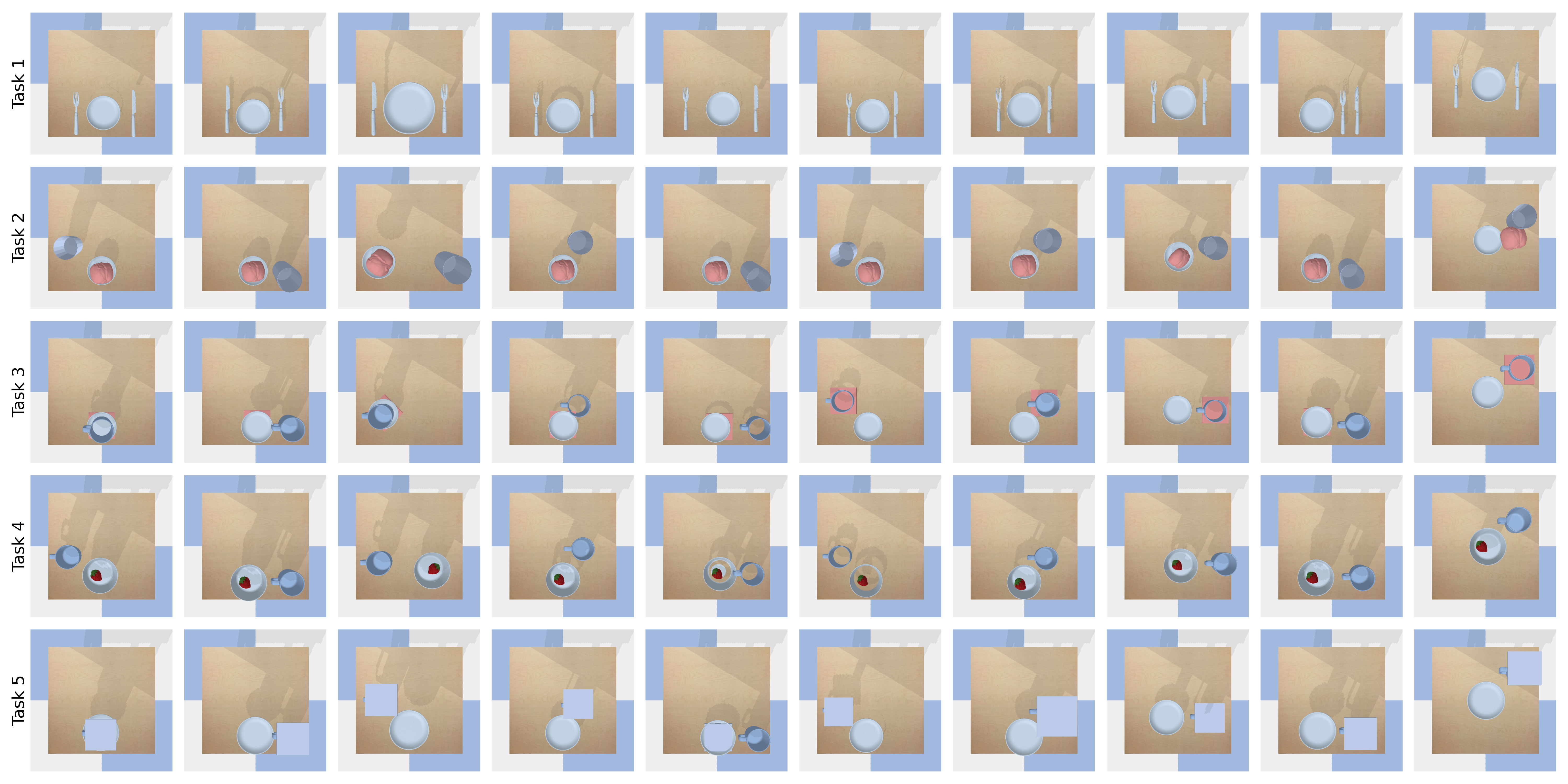}
  \caption{\textbf{\texttt{LLM-GROP} task user preferences.} Figure shows differing preferences for 10 users on train sub-tasks.}
  \label{fig:llmg_rep}
\end{figure*}

\subsubsection{Additional experiments.} Finally, we also investigate if \algoname\ is susceptible to performance degradation if the order of demonstrations is altered. Our experiments showed that the effect of re-ordering diminishes after about 12 samples which shows the robustness of \algoname. We also deploy \algoname\ on a mobile robot and after some optimizations (\eg inclusion enumeration, batched pixel inference, caching the outputs) it runs approximately at 1 Hz for a single GPU single process execution. Details of these additional experiments are included in the appendix.

\end{document}